# The Algonauts Project 2021 Challenge:
## *How the Human Brain Makes Sense of a World in Motion*

**Radoslaw Martin Cichy**[1](radoslaw.cichy@fu-berlin.de), **Kshitij Dwivedi**[1,2](kshitijdwivedi93@gmail.com), **Benjamin Lahner**[3](blahner@mit.edu), **Alex Lascelles**[3](alexlasc@mit.edu), **Polina Iamshchinina** (iampolina@zedat.fu-berlin.de) [1], **Monika Graumann**[1](monikag@zedat.fu-berlin.de), **Alex Andonian**[3](andonian@mit.edu), **N Apurva Ratan Murty**[4](ratan@mit.edu) **Kendrick Kay**[5](kay@umn.edu), **Gemma Roig**[2](roig@cs.uni-frankfurt.de)*, **Aude Oliva**[3](oliva@mit.edu)*

[1] Department of Education and Psychology, Freie Universität Berlin, Berlin, Germany
[2] Department of Computer Science, Universität Frankfurt, Frankfurt, Germany
[3] Computer Science and Artificial Intelligence Laboratory, MIT, Cambridge, USA
[4] Department of Brain and Cognitive Science, MIT, Cambridge, USA
[5] Center for Magnetic Resonance Research (CMRR), Department of Radiology, University of Minnesota, Minneapolis, Minnesota, USA

## Abstract:

**The sciences of natural and artificial intelligence are fundamentally connected. Brain-inspired human-engineered AI are now the standard for predicting human brain responses during vision, and conversely, the brain continues to inspire invention in AI. To promote even deeper connections between these fields, we here release the 2021 edition of the Algonauts Project Challenge:** *How the Human Brain Makes Sense of a World in Motion* **(http://algonauts.csail.mit.edu/). We provide whole-brain fMRI responses recorded while 10 human participants viewed a rich set of over 1,000 short video clips depicting everyday events. The goal of the challenge is to accurately predict brain responses to these video clips. The format of our challenge ensures rapid development, makes results directly comparable and transparent, and is open to all. In this way it facilitates interdisciplinary collaboration towards a common goal of understanding visual intelligence. The 2021 Algonauts Project is conducted in collaboration with the Cognitive Computational Neuroscience (CCN) conference.**

**Keywords: artificial intelligence; human neuroscience; vision; event understanding; prediction; challenge; benchmark**

## Introduction

**The 2021 Edition of the Algonauts Project: How the Human Brain Makes Sense of a World in Motion**

This 2021 edition is the second edition of the Algonauts Project challenge and is titled "How the Human Brain Makes Sense of a World in Motion". Its goal is to determine which computational model best explains human brain responses while humans view everyday events.

We focus on event understanding from videos because it is of current interest to the sciences of biological and artificial intelligence alike: it is a fundamental cognitive capacity enabling intelligent behavior, and it is an unsolved problem and hot topic in computer vision (Monfort et al., 2020).

**Related prediction challenges in neuroscience**
The 2021 edition continues the spirit of the 2019 edition of Algonauts in its goal of explaining human visual brain responses (Cichy et al., 2019). But it also goes beyond the 2019 challenge in that it **a)** focuses on videos rather than still images, **b)** provides a richer stimulus set and **c)** assesses responses across the whole brain.

It further relates to initiatives such as the "The neural prediction challenge" at Berkeley and "brain-score" (http://www.brain-score.org/) (Schrimpf et al., 2020) which also establish benchmarks and leaderboards. The Algonauts Project 2021 differs from these complementary efforts by emphasizing human data, by focusing on videos rather than still images, by coupling neural prediction benchmarks to a challenge limited in time, and by the educational and collaborative components that will be incorporated as a dedicated session at the Cognitive Computational Neuroscience (CCN) conference in 2021.

## Materials and Methods for the 2021 Algonauts Challenge

The target of the challenge is to predict human brain responses while participants view short video clips of everyday events (e.g., panda eating, fish swimming, a person paddling).

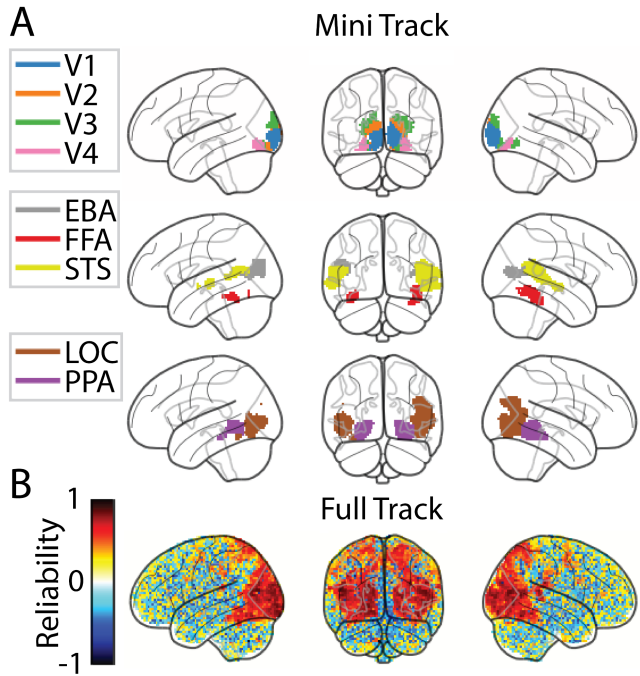

**Figure 1. Challenge tracks.** **A) Mini track**: we consider brain responses in a set of key regions of the visual brain, starting with early and mid-level visual cortex (*V1, V2, V3,* and V4) and extending into higher-level cortex that responds preferentially to all objects or particular categories (Body- EBA; Face - FFA, STS; Object - LOC; Scene - PPA), areas known to elicit high neural activity in humans viewing natural stimuli. **B) Full track**: we consider all voxels in the brain that show reliable responses to video clips (hot colors). Both panels are for a representative subject.

For the challenge we provide brain data measured with functional magnetic resonance imaging (fMRI). fMRI is a widely used brain imaging technique with high spatial resolution that measures blood flow changes associated with neural responses. We measured fMRI at 3T in 2.5×2.5×2.5mm cubes (called voxels) while 10 human participants viewed a set of 1,102 3-s long video clips of everyday events. 1,000 of these videos were repeated 3 times and form the training set of the challenge. 102 videos were repeated 10 times and form the testing set.

After performing basic fMRI pre-processing (realignment, coregistration), we estimated a single activation value for each video and voxel in the brain (for further details see here: link). We then split the data into training and test data sets.

**Training Data Set.** 1,000 video clips and the associated brain data form the training data set. These data can be used to train the parameters of a computational model and/or how it might map onto brain responses.

**Test Data Set.** The test set consists of 102 short videos and the respective brain responses. Participants in the challenge receive only the test videos; the brain data is held back. Participants submit predicted brain responses from their models (i.e., synthetic brain data), and we evaluate how good the prediction is using the held-out brain data.

**Challenge Definition.** We have created two related but independent tracks in the challenge. The **Mini Track** (**Fig. 1A**) focuses on responses in a core set of 9 cortical regions in the visual brain. The **Full Track** (**Fig. 1B**) takes a spatially unbiased perspective and includes voxels at all brain locations at which responses were reliably detected.

Participants are required to predict voxel responses to the test videos **(Fig. 2A)**.

Participants submit the predicted synthetic brain data, which is compared against the held-out brain data (**Fig 2B**). The comparison metric for both tracks is how well predicted synthetic brain responses resemble empirically recorded brain responses. In detail, we correlate the predicted brain data with the empirical data of each voxel on the test videos and then average the correlation across all the voxels in consideration. Aggregated across visual brain regions (**Mini Track**) or the whole brain (**Full Track**), this yields a challenge score that determines the relative place in the leaderboard.

Participants are free to predict brain responses from the stimulus videos in any way they see fit (see rules). As an example, we guide the reader through one common approach called a voxel-wise encoding model (Wu et al., 2006; Naselaris et al., 2011) where the response of each voxel is predicted independently using the multiple features provided by a computational model (a regularized linear regression is typically used to form the prediction). An example implementation is included in the development kit.

**Voxel-wise encoding model: an example approach**

In a voxel-wise encoding model approach, the prediction of brain responses typically has two steps (**Fig. 2C**). In the first step, activations of a computer vision model to videos are extracted. This changes the format of the data (from pixels to model features) and typically reduces the dimensionality of the data. The features of a given model are interpreted as a potential hypothesis about the features that a given brain area might be using to represent the stimulus.

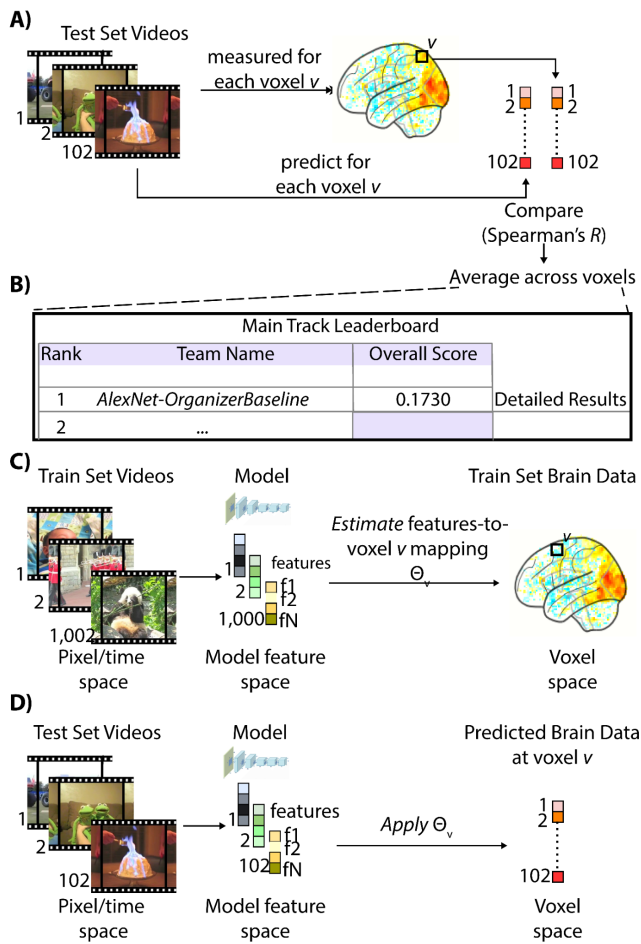

**Figure 2. Goal and example approach for the Algonauts 2021 edition challenge.** **A)** The overall goal of the challenge is to predict brain responses to videos measured with fMRI (i.e., voxel $v$'s responses) during video stimulus presentation. **B)** Determining model performance. For each voxel we correlate its predicted response with the empirical held-back responses to the test videos. The averaged correlation (across voxels or regions) results in a score that determines the relative place in the leaderboard. **C)** First step of voxel-wise encoding model procedure. We extract the activations of a computer vision model to training videos, and estimate a mapping $\phi$ between model activations to a given voxel $v$'s responses using the training data. Typically, the mapping is linear and is estimated using regularized regression. **D)** Second step of voxel-wise encoding model procedure. We extract the activations of the computational model to the test videos, and then apply the estimated mapping $\phi$ to predict voxel $v$'s response on the test data.

In the second step **(Fig. 2D)**, activations of the computational model are linearly mapped onto each voxel's responses. This step is necessary as there is not necessarily a one-to-one mapping between voxels and model features. Instead, each voxel's response is hypothesized to correspond to a weighted combination of activations of multiple features of the model. If the computational model is a suitable model of the brain, the mapped predictions will fit empirically recorded data well.

We provide an example implementation of the voxel-wise encoding model using AlexNet (Krizhevsky et al., 2012) as the computational model in the development kit.

**Development Kit.** The development kit contains a set of Python scripts to facilitate participation in the challenge. It provides example code to **a)** extract model activation values to the videos from a deep neural network (i.e., Alexnet), **b)** estimate a mapping between model activations and brain responses (i.e., a voxel-wise encoding model), **c)** evaluate the predictions using cross-validation on the training data in the absence of testing data, **d)** visualize the results in brain space, and **e)** prepare predicted brain responses required for submission to the challenge website.

**Baseline Model.** We evaluate AlexNet (Krizhevsky et al., 2012) *as* a typical example of a deep neural network that predicts human brain responses to still images well. The noise-normalized predictive value of AlexNet is In the mini track (regions) $R = 0.3673$ and in the full track (whole brain) $R = 0.1730$.

**Rules.** To encourage broad participation the challenge has a simple submission process with minimal rules. Participants can use any model derived from any source and trained on any type of data, and can use any method to map model activations to predict brain responses. However, we explicitly forbid the use of primate brain responses to the test video set. We request participants to eventually submit a short report to a preprint server, making their reasoning of their specific model and the procedure to predict brain responses explicit.

## Discussion

**Open challenges as scientific instruments in cognitive science.** Considered as scientific instruments, open challenges have a uniquely advantageous set of properties for topics at the intersection of artificial and natural intelligence.

For one, by their very nature open challenges enforce transparency and openness. These are values recognized to promote replicability of results (Nosek et al., 2015; Poldrack et al., 2017) that may conflict with historically grown research and publishing practices.

Second, quantitative benchmarks ensure a precise metric of success. As testing a model amounts to testing a hypothesis about brain function, this contributes to theory building.

Third, challenges on topics that are addressed in different scientific fields with disparate success criteria (e.g., theoretical explanation of natural intelligence, engineering artificial intelligence) can foster communication and collaboration by providing a common platform with a common goal. The sciences creating artificial intelligence can be guided and inspired by the human brain, as the brain is a feasibility demonstration of a robust, highly efficient and low-energy solution to many problems in AI.

Finally, the scientific instrument of challenges enables collaboration at an intermediate (meso-scale) level of complexity, i.e. when more than single laboratories on their own (micro-scale), but less than scientific institutes with a unified goal (macro-scale) are most likely to efficiently and swiftly drive a field forward. We believe that many problems at the intersection of the sciences of artificial and natural intelligence are at that level.

**Prediction vs. explanation.** Predictions and explanations are not interchangeable, but are closely related and complement each other. For one, explanations can only be ultimately full and successful if they can also provide successful predictions (Breiman, 2001; Yarkoni and Westfall, 2017). Further, inspection of what made a model successful can provide very useful information in at least two ways. It can help to select between candidate models for further inquiry that are indistinguishable on theoretical grounds only, and it can guide attention to investigation of the properties that make particular models successful. Finally, predictive success as an evaluation criterion circumvents the challenges of evaluation on purely theoretical grounds.

This is not to mean, again, that explanation and interpretation are not or less important. Our rules explicitly exclude the use of brain responses recorded to the test set, because one brain can be a very good predictor of another brain, but we would not learn much about the brain by substituting one brain for another.

**The next edition of the project**. A recognized need among researchers of artificial and biological intelligence is brain data at larger scale: to fit or build more powerful computational models more data is needed. In the next version of the Algonauts Project we plan to use the Natural Scenes Dataset (Allen et al., 2021), which provides high-resolution 7T fMRI brain responses for more than 70,000 still images. Parts of the NSD data set are being held back to be used as testing data for the next Algonauts Challenge. This will allow not only the assessment of computational models in predicting brain responses with unprecedented precision, but also drive the training/machine learning of models to be informed by rich brain responses.

## Acknowledgments

This research was funded by DFG (CI-241/1-1, CI-241/1-3,CI-241/1-7) and ERC grant (ERC-2018-StG) to RMC; NSF award (1532591) in Neural and Cognitive Systems and the Vannevar Bush Faculty Fellowship program funded by the ONR (N00014-16-1-3116) to A.O; the Alfons and Gertrud Kassel foundation to G.R. We also thank the MIT-IBM Watson AI Lab for support. The experiments were conducted at the Athinoula A. Martinos Imaging Center at the McGovern Institute for Brain Research, Massachusetts Institute of Technology, on a Siemens PrismaFit 3T scanner (Erlangen, Germany) supported with funding from a NIH Shared Instrumentation Grant (1S10OD021569).